\documentclass{sig-alternate}
\pdfoutput=1
\usepackage{amsfonts}
\usepackage{amssymb}
\usepackage{graphicx}
\usepackage{subfigure}
\usepackage{epstopdf}
\usepackage[ruled]{algorithm2e}

\usepackage{microtype}

\newtheorem{thm:def}{Definition}
\newtheorem{thm:lem}{Lemma}
\newcommand{\mU}{{\mathbf{U}}}

\newcommand{\mV}{{\mathbf{V}}}
\newcommand{\mP}{{\mathbf{P}}}
\newcommand{\dmP}{{\Delta\mP}}
\newcommand{\mQ}{{\mathbf{Q}}}
\newcommand{\mY}{{\mathbf{Y}}}

\newcommand{\hmY}{{\hat{\mY}}}
\newcommand{\vx}{{\mathbf{X}}}
\newcommand{\vz}{{\mathbf{Z}}}
\newcommand{\rR}{{\mathbb{R}}}
\newcommand{\trs}{{T}}

\newcommand{\upL}{\tilde{L}}
\newcommand{\upPL}{\tilde{L}^p}

\newcommand{\Idx}{\mathcal{O}}
\newcommand{\vzero}{\mathbf{0}}

\newcommand{\PCD}[1]{PL2M-{#1}}
\newcommand{\SCD}[1][ ]{L2M{#1}}
\begin{document}
\title{A Parallel and Efficient Algorithm for Learning to Match}
\numberofauthors{1}
\author{
 \alignauthor
 Jingbo Shang$^{1,4}$\titlenote{},\hspace{2mm}
 \global\titlenotecount=0
 Tianqi Chen$^2$\titlenote{These authors contributed equally to this work. The short version of this paper~\cite{ICDM-short} was published in ICDM 2014.\vspace{-10mm}},
 \hspace{2mm} Hang Li$^3$,\hspace{2mm} Zhengdong Lu$^3$,\hspace{2mm} Yong Yu$^4$
  \and
  \alignauthor
  \affaddr{$^1$University of Illinois at Urbana Champaign, IL, USA}\\
  \email{shang7@illinois.edu}
  \and
  \alignauthor
  \affaddr{$^2$University of Washington, Seattle, WA, USA}\\
  \email{tqchen@cs.washington.edu}
  \and
  \affaddr{$^3$Huawei Noah's Ark Lab, Hong Kong}\\
  \email{\{hangli.hl, lu.zhengdong\}@huawei.com}
  \and
  \alignauthor
  \affaddr{$^4$Shanghai Jiao Tong University, Shanghai, China}\\
  \email{\{shangjingbo, yyu\}@apex.sjtu.edu.cn}
}

\maketitle
\begin{abstract}
Many tasks in data mining and related fields can be formalized as matching between objects in two heterogeneous domains, including collaborative filtering, link prediction, image tagging, and web search. Machine learning techniques, referred to as learning-to-match in this paper, have been successfully applied to the problems. Among them, a class of state-of-the-art methods, named feature-based matrix factorization, formalize the task as an extension to matrix factorization by incorporating auxiliary features into the model. Unfortunately, making those algorithms scale to real world problems is challenging, and simple parallelization strategies fail due to the complex cross talking patterns between sub-tasks. In this paper, we tackle this challenge with a novel parallel and efficient algorithm for feature-based matrix factorization. Our algorithm, based on coordinate descent, can easily handle hundreds of millions of instances and features on a single machine. The key recipe of this algorithm is an iterative relaxation of the objective to facilitate parallel updates of parameters, with guaranteed convergence on minimizing the original objective function. Experimental results demonstrate that the proposed method is effective on a wide range of matching problems, with efficiency significantly improved upon the baselines while accuracy retained unchanged.
\end{abstract}





\section{Introduction}
Many application tasks can be formalized as matching between objects in two heterogeneous domains, in which the association between some objects and information on those objects are given. We refer to the objects from one domain as queries and those from the other as targets, with the distinction usually clear from the context. For example, in collaborative filtering, given some items, one manages to find the users who have best match to the items, by using the preference of some users on some items as well as the features of users and items. Another example is image tagging, in which one wants to associate tags (keywords) with images based on some tagged images as well as the features of tags and images. Recent years have observed a great success of employing machine learning techniques, referred to as learning-to-match in this paper, to solve the matching problems.

Among existing approaches, a family of factorization models that make use of feature spaces to encode additional information, stand out as state-of-the-art in matching tasks. Examples include factorization machines~\cite{Rendle:ICDM10, rendle:tist2012}, feature-based latent factor models for link prediction~\cite{Chen:SIGIR2012, Menon:ECML11}, and regression-based latent factor models~\cite{agarwal:kdd}.
We refer to this class of methods as feature-based matrix factorization (FMF) in this paper. The basic idea of FMF is to formalize the task as extension to plain matrix factorization for incorporating the features of objects into the model. In this way, one can make a full use of available information in the task to improve the accuracies. In fact, FMF is the best performer on many real world matching tasks. In collaborative filtering, FMF models using user feedback~\cite{Koren:KDD08, Koren:KDD09}, attribute~\cite{agarwal:kdd,stern:www}, and content~\cite{Chen:SIGIR2012, Weston:ICML12} have outperformed other models including plain matrix factorization. In web search, FMF models for calculating matching scores (relevance) between queries and documents have significantly enhanced relevance ranking~\cite{Wu:WSDM13, wu2013learning}. FMF models have also been successfully employed in link prediction~\cite{Menon:ECML11}, and have been adopted by the champion teams in KDD Cup 2012~\cite{SVDFeature:JMLR, rendle:tist2012}.

The learning of the FMF model can be conducted with a coordinate descent algorithm or a stochastic gradient descent algorithm. Since a matching problem is usually of a very large scale, with hundreds of millions of objects and features or more, it can easily become hard for FMF to manage. It is therefore necessary to develop a parallel and efficient algorithm for FMF. This is exactly the problem we attempt to address in this paper.

Making FMF scalable and efficient is much more difficult than it appears, due to the following two challenges. First, training requires simultaneous access to all the features, and thus the existing techniques for parallelization of matrix factorization~\cite{Gemulla:KDD11, Yu:ICDM12, Zhuang:Recsys2013} are not directly applicable. Second, the computation complexity of the coordinate descent algorithm is still too high, and it can easily fail to run on a single machine when the scale of problem becomes large, calling for techniques to significantly accelerate the computation. By making use of repeating patterns, the least-squares and probit losses can be scaled up for coordinate descent~\cite{rendle2013scaling}, but it does not provide guarantee for any general convex loss functions. Existed parallel coordinate descent algorithms, such as~\cite{collins2002logistic}~and~\cite{mukherjee2013parallel}, due to the complex feature dependencies, cannot be directly applied here. The Hogwild!~\cite{hogwild} algorithm for parallel stochastic gradient descent can be applied here, but it is a generic algorithm and thus is still inefficient for FMF.

In this paper, we try to tackle the two challenges by developing a parallel and efficient algorithm tailored for learning-to-match. The algorithm, referred to as parallel and efficient algorithm for learning-to-match (\textsc{PL2M}), parallelizes and accelerates the coordinate descent algorithm through (1) iteratively relaxing the objective to facilitate parallel updates of parameters, and (2) avoiding repeated calculations caused by features. The main contributions of this paper are as follows.
\begin{itemize}
\item We propose the parallel and efficient algorithm for feature-based matrix factorization, which iteratively relaxes the objective for parallel updates of parameters, and neatly avoids repeated calculations caused by features, for any general convex loss functions.
\item We theoretically prove the convergence of the proposed algorithms on minimizing the original objective function, which is further verified by our extensive experiments. The parallel algorithm can automatically adjust the rate of parallel updates according to the conditions in learning.
\item We empirically demonstrate the effectiveness and efficiency of the proposed algorithm on four benchmark datasets. The parallel algorithm achieves nearly linear speedup and the proposed acceleration helps the parallel algorithm run about $5$ times faster than the Hogwild!~\cite{hogwild} algorithm on average, using $8$ threads.
\end{itemize}
Given the importance of the FMF models and difficulty of their parallelization, the work in this paper represents a significant contribution to the study of learning to match. To our best knowledge, this is the first effort on the scalability of the general FMF models.

The rest of the paper is organized as follows. Section~\ref{sec:l2m} gives a formal description of the generalized matrix factorization and Section~\ref{sec:ecd} explains the efficient coordinate descent algorithm. Section~\ref{sec:parallel-update} describes parallelization of the coordinate descent algorithm. Related work is introduced in Section~\ref{sec:rel}. Experimental results are provided in Section~\ref{sec:exp}. Finally, the paper is concluded in Section~\ref{sec:con}.

\section{Learning to Match}\label{sec:l2m}
In this section, we give a formal definition of learning-to-match and a formulation of a feature-based matrix factorization. We also present our motivation of parallelizing this learning task.

\subsection{Problem Formulation}
Learning-to-match can be formally defined as follows. Let $\vx=[ \vx_1, \vx_2 \cdots \vx_q ]$ be the instances in the query domain and $\vz = [\vz_1, \vz_2 \cdots \vz_p ]$ be the instances in the target domain, where $\vx_i \in \rR^{n \times 1}$ and $ \vz_j \in \rR^{m \times 1}$ are query and target instances (feature vectors) respectively. For some query-target pairs, the corresponding matching scores $\{\mY_{ij} | (i,j) \in \Idx \}$ are given as training data, where $\Idx$ is the set of indices for all observed query-target pairs. Our problem is to learn to predict the matching score $\hmY_{ij} = f( \vx_i, \vz_j )$ between any pair of query $i$ and target $j$.

The setting is rather general and it subsumes many application problems. For example, in collaborative filtering, a user's preference over an item can be interpreted as the matching score between the user and the item. In social link prediction, the likelihood of link between nodes on the network can be as regarded as the matching score between the nodes. Web search, in general document retrieval, can also be formalized as a problem of first matching between a given query and documents and then ranking of documents based on the matching scores.

The goal of learning-to-match is to make accurate prediction by effectively using the information on the given relations between instances (e.g., similar users may prefer similar items), as well as the information on the features of instances (e.g., users may prefer items with similar properties).

\subsection{Model}
\begin{figure}[t]
\centering
    \includegraphics[scale=1]{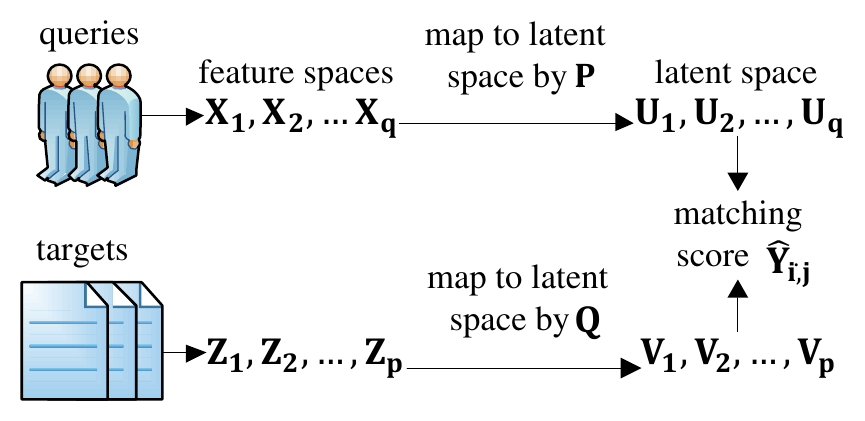}
    \caption{The general feature-based matrix factorization model for learning-to-match, which we have accelerated and parallelized in this paper.}
    \label{fig:fmf}
\end{figure}

The query and target instances (feature vectors) are in two heterogeneous feature spaces, and a direct match between them is generally impossible. Instead, we map the feature vectors in the two domains into a latent space and perform matching on the images of the feature vectors in the latent space. We calculate the matching score of a query-target pair as
\begin{equation}
    \hmY_{ij} = \mU_i^T\mV_j,\ \mU_i = \mP\vx_i, \mV_j = \mQ\vz_j
    \label{eq:model}
\end{equation}
where $\mP \in \rR^{d \times n}$ and $\mQ \in \rR^{d \times m}$ are transformation matrices that map feature vectors from the feature spaces into the latent space. $\mU_i$ and $\mV_j$ are latent factors of query instance $i$ and target instance $j$. In this paper, we use $\mU_i$ to denote the $i$th column of matrix $\mU$ and $\mV_j$ to denote the $j$th column of matrix $\mV$. We refer to the model in Equation~(\ref{eq:model}) as the model of feature-based matrix factorization. The model can be also interpreted as linear matching function of latent factors, in which the latent factor of each instance is also linearly constructed from feature vectors. The query latent factor $\mU_i$ can be expressed as $\mU_i = \sum_{s = 1}^{n} \vx_{is} \mP_s$. The target latent factor $\mV_j$ can be expressed similarly.

The model, as shown in Figure~\ref{fig:fmf}, contains many existing models of feature-based matrix factorization as special cases~\cite{Rendle:ICDM10, rendle:tist2012, Chen:SIGIR2012, Menon:ECML11, agarwal:kdd}.
When no ``informative" features are available for objects of both domains, the feature matrices contain only the indices of the objects. Clearly, in such cases $\vx$ and $\vz$ become identity matrices of sizes $q\times q$ and $p\times p$, the feature-based matrix factorization model naturally degenerates to the plain matrix factorization~\cite{Koren:MF09}.

The objective of the learning task then becomes~\footnote{$\mU$ and $\mV$ are not model parameters; they are only auxiliary variables determined by $\mP$ and $\mQ$.}:
\begin{equation}
    \mP, \mQ = \mbox{argmin}_{\mP, \mQ} \sum_{i,j\in \Idx} l(\hmY_{ij}, \mY_{ij} ) + \Omega( \mP ) +  \Omega( \mQ ) \label{eq:obj}
\end{equation}

Here $l(\hmY_{ij}, \mY_{ij})$ is a strongly convex loss function that measures the difference between the prediction $\hmY_{ij}$ and the ground truth $\mY_{ij}$. The loss function $l$ can be square loss for regression
\begin{equation}
    l(\hmY_{ij}, \mY_{ij}) = (\hmY_{ij} - \mY_{ij}) ^ 2
\end{equation}
or logistic loss for classification
\begin{equation}
    l(\hmY_{ij}, \mY_{ij}) = \mY_{ij} \ln(1 + e^{-\hmY_{ij}}) + (1 - \mY_{ij}) \ln(1 + e^{\mY_{ij}})
\end{equation}
where $\mY_{ij} \in [0, 1]$. $\Omega$ is a regularization term based on elastic net~\cite{Zou:ElasticNet2005} which includes $l_2$ (ridge) and $l_1$ (lasso) regularization as its special cases.
\begin{equation}
    \Omega( \mP )= \alpha || \mP ||_1 + \frac{\lambda}{2} ||\mP||_2^2,\ \Omega( \mQ )= \alpha || \mQ ||_1 + \frac{\lambda}{2} ||\mQ||_2^2
\end{equation}


\SetAlgoSkip{}
\begin{algorithm}[t]
    \caption{Coordinate Descent Algorithm for Learning to Match}\label{alg:cd-old}
    $\mbox{randomly initialize }\mP,\mQ$\;
    $\mU \leftarrow \mP \vx$, $\mV \leftarrow \mQ \vz$,  $\hmY \leftarrow \mU^\trs \mV$ \{calculate $\mU$, $\mV$, $\hmY$\}\\
    \While{not converge}{
        \For{$k=1$ {\bfseries to} $d$}{
            \For{$s=1$ {\bfseries to} $n$}{
                $x \leftarrow \sum_{i,j\in \Idx} g_{ij} \mV_{kj} \vx_{is}$ \\
                $y \leftarrow \beta\sum_{i,j\in \Idx} \mV_{kj}^2 \vx_{is}^2$ \\
                $\dmP_{ks} \leftarrow  T ( x, y, \mP_{ks}, \alpha, \lambda )$\\
                $\mP_{ks} \leftarrow \mP_{ks} + \dmP_{ks} $\\
            }
        }
        $\mU = \mP \vx$  \{recalculate buffered $\mU$\}\\
        update $\mQ$ in the same way as $\mP$\\
    }
\end{algorithm}

\subsection{Performance}
The use of features in learning-to-match is crucial for the accuracy of the task. Usually an FMF model, when properly optimized, can produce a higher accuracy in prediction than a model of plan matrix factorization (MF). This is because the FMF model can leverage more information for the prediction, particularly the feature information, while the MF model can only rely on relations between instances which are usually very sparse. For example, in the task of recommendation~\cite{Chen:SIGIR2012}, only about $0.1\%$ entries are observed. In Tencent Weibo dataset at KDD Cup 2012, about $70\%$ of users in the test set have no following records in the training set~\cite{Chen:KDDCUP}. As a result, MF cannot achieve satisfactory results in the tasks, while FMF models give the best results.

In fact, it has been observed the FMF models (with different types of features used) achieve state-of-the-art results on many different tasks, outperforming the models of MF with big margin. For example, in collaborative filtering, user feedback (SVD++)~\cite{Koren:KDD08}, user attribute~\cite{agarwal:kdd}, and product attribute~\cite{stern:www} are incorporated into models to further improve the accuracies in prediction. In web search~\cite{Wu:WSDM13, wu2013learning}, term vectors of queries and documents are used as features to significantly improve relevance ranking. FMF models also give the best results in link prediction in KDD Cup 2012~\cite{Menon:ECML11, Chen:SIGIR2012, Chen:KDDCUP}.

\SetAlgoSkip{}
\begin{algorithm}[t]
    \caption{Efficient Algorithm for Learning to Match}\label{alg:cd-group}
    $\mbox{randomly initialize }\mP,\mQ$\\
    $\mU \leftarrow \mP \vx$, $\mV \leftarrow \mQ \vz$,  $\hmY \leftarrow \mU^\trs \mV$ \{calculate $\mU$, $\mV$, $\hmY$\}\\
    \While{not converge}{
        \For{$k=1$ {\bfseries to} $d$}{
            \For{$i=1$ {\bfseries to} $q$}{
                $G_{ki} \leftarrow \sum_{j \in \Idx_i }  g_{ij} \mV_{kj}$ \\
                $H_{ki} \leftarrow \beta \sum_{j \in \Idx_i}\mV^2_{kj}$\\
            }
            \For{$s=1$ {\bfseries to} $n$}{
                $x \leftarrow \sum_{i} G_{ki} \vx_{is} $ \\
                $y \leftarrow \sum_{i} H_{ki} \vx_{is}^2$ \\
                $\dmP_{ks} \leftarrow  T ( x, y, \mP_{ks}, \alpha, \lambda )$\\
                $\mP_{ks} \leftarrow \mP_{ks} + \dmP_{ks} $\\
                \For{$i=1$ {\bfseries to} $q$}{
                    $G_{ki} \leftarrow G_{ki} +  \vx_{is}\dmP_{ks} H_{ki} $\\
                }
            }
        }
        $\mU = \mP \vx$ \{recalculate buffered $\mU$\}\\
        update $\mQ$ in the same way as $\mP$\\
    }
\end{algorithm}

\subsection{Scalability}

The success of the FMF models strongly indicates the necessity of scaling up the corresponding learning algorithms, given that the existing algorithms still cannot easily handle large datasets. By making use of repeating patterns, the least-squares and probit losses can be scaled up for coordinate descent~\cite{rendle2013scaling}, but it does not guarantee for any general convex loss functions. Other algorithms of parallel coordinate descent, such as~\cite{collins2002logistic}~and~\cite{mukherjee2013parallel} cannot be directly applied to FMF, because it is difficult for them to handle complex feature dependencies in FMF. The Hogwild!~\cite{hogwild} algorithm for parallel stochastic gradient descent can be applied here, but it is a generic algorithm and thus is still inefficient for FMF. To our best knowledge, our work in this paper is the first effort on scalability of learning-to-match, i.e., feature-based matrix factorization.

\section{Efficient Algorithm for Learning to Match}\label{sec:ecd}
In this section, we propose an acceleration of the coordinate descent algorithm for solving the feature-based matrix factorization problem. We prove the convergence of the accelerated algorithm, and we also give its time complexity at Section~\ref{sec:timeComplextiy}.

\subsection{Coordinate Descent Algorithm}
Let $g_{ij} = \frac{\partial}{\partial \hmY_{ij}} l( \hmY_{ij}, \mY_{ij} )$ be the gradient of each instance over prediction and $\beta$ be constant of $l$, such that,
\begin{equation}
    l(y + dy) \leq l(y) + l'(y) dy + \frac{\beta}{2} dy^2
\end{equation}
That is, $\beta = \sup_{y} \frac{\partial^2}{\partial y^2} l( y, \mY_{ij} )$.
We can exploit the standard technique to learn the model using coordinate descent (CD), as shown in Algorithm~\ref{alg:cd-old}\footnote{In this paper, all matrix operations in algorithms have taken the advantages of sparsity, e.g. summations are all over nonzero entries implicitly. So do the time complex analysis and implementations of all algorithms.}. Here, $T$ is defined using the following thresholding function to handle optimization of $l_1$ norm
\begin{equation}
  T ( x, y, w, \alpha, \lambda ) = \left\{
  \begin{array}{ll}
   \max( -\frac{x+\lambda w + \alpha}{y+\lambda}, - w ) & w -\frac{x+\lambda w}{y+\lambda} \geq 0 \\
   \min( -\frac{x+\lambda w - \alpha}{y+\lambda}, - w ) & w -\frac{x+\lambda w}{y+\lambda} < 0 \\
  \end{array} \right.
\end{equation}

The regularization term only affects the result through function $T$, and thus makes most part of the algorithm independent of regularization. Note that we implicitly assume $\hmY_{ij}$ is buffered and kept up to date, when $g_{ij}$ is needed in the algorithm.

The time complexity of one update in Algorithm~\ref{alg:cd-old} is $O( d |\Idx|( \frac{\|\vx\|_0}{q}  + \frac{\|\vz\|_0}{p} ) )$,  where $\|\vx\|_0$ and $\|\vz\|_0$ denote numbers of nonzero entries in feature matrices $\vx$ and $\vz$, $q$ and $p$ denote numbers of query and target instances, and $(\frac{\|\vx\|_0}{q}  + \frac{\|\vz\|_0}{p})$ denotes average number of nonzero features for each pair $(i,j)\in \mathcal{O}$. We note that the time complexity is the same as the time complexity of stochastic gradient optimization for Equation ~(\ref{eq:obj}). From the analysis, we can see that the time complexity of algorithm is increased by order of $O( d|\Idx| )$ when average number of nonzero features increases. This can greatly hamper the learning of matching model, when a large number of features are used.

\subsection{Acceleration}
We give an efficient algorithm for learning-to-match by avoiding the repeated calculations caused by features. There is only a little works focused on the acceleration of the FMF models. The most relevant one is that scaling up some specific coordinate descent by making use of repeating patterns of features~\cite{rendle2013scaling}. However, it is specialized for the least-squares and probit losses. Although the idea is similar to the avoiding repeated calculations in our work, we extend the idea to any general convex loss functions.

One can see that there exist repeated calculations of summations for the same query or target when calculating $x$ and $y$ in Algorithm~\ref{alg:cd-old}, which gives us a chance to speed up the algorithm. We introduce two auxiliary variables $G_{ki}$ and $H_{ki}$ calculated by
\begin{equation}
    G_{ki} = \sum_{j \in \Idx_i }  g_{ij} \mV_{kj},\ H_{ki} =\beta \sum_{j \in \Idx_i }   \mV^2_{kj}
\end{equation}
where $\Idx_i=\{j|(i,j)\in \Idx\}$ is the set of observed target instances associated with query instance $i$. The key idea of efficient CD is to make use of $G_{ki}$ and $H_{ki}$ to save duplicated summations in Algorithm~\ref{alg:cd-old}. Since the gradient value $g_{ij}$ is changed after each update, it is not trivial to let $G_{ki}$ unchanged. Our algorithm keeps making $G_{ki}$ updated to ensure the convergence of the algorithm. The efficient algorithm for learning-to-match is shown in Algorithm~\ref{alg:cd-group}.

\subsection{Convergence of Algorithm}
Next, we prove the convergence of Algorithm~\ref{alg:cd-group}, which greatly reduces the time complexity of learning.
Suppose that the $k$th row of $\mP$ is changed by $\dmP_{k,:}$. After the change, the loss function can be bounded by
\begin{equation}
\begin{split}
L( \dmP_{k,:} ) =& \sum_{i,j\in \Idx} l( \hmY_{ij} + (\sum_{s} \dmP_{ks} \vx_{is}) \mV_{kj}, \mY_{ij} ) \\
\leq & \sum_{i,j\in \Idx} l( \hmY_{ij}, \mY_{ij} ) +  \sum_{i,j\in \Idx} g_{ij} \mV_{kj} (\sum_{s} \dmP_{ks} \vx_{is}) \\
& + \frac{1}{2} \sum_{i,j\in \Idx} h_{ij} \mV_{kj}^2 (\sum_{s} \dmP_{ks} \vx_{is})^2 \\
=& \sum_{i,j\in \Idx} l( \hmY_{ij}, \mY_{ij} ) + \sum_{i} G_{ki} (\sum_{s} \dmP_{ks} \vx_{is})\\
& + \frac{1}{2} \sum_{i} H_{ki}(\sum_{s} \dmP_{ks} \vx_{is})^2 \\
\triangleq& \upL( \dmP_{k,:} )
\end{split}
\end{equation}
Intuitively, updating $G_{ki}$ corresponds to minimizing the quadratic upper bound $\upL(\dmP_{k,:})+\Omega( \mP_{k,:}+\dmP_{k,:} )$ of the original convex loss which is re-estimated each round.
Formally, all the values of $\dmP_{k,:}$ are zero in the beginning. We need to sequentially update $\mP_{ks}$ for different $s$'s to minimize $\upL(\dmP_{k,:})+\Omega( \mP_{k,:}+\dmP_{k,:} )$. Assuming that we have already updated $\mP_{k,1}$ $\cdots$ $\mP_{k,t-1}$ and need to decide $\dmP_{kt}$, we can calculate the upper bound $\upL$ as follows
\begin{equation}
\begin{split}
\upL =& c + \sum_{i} \left( G_{ki} + \sum_{s=1}^{t-1} \vx_{is} \dmP_{ks} H_{ki} \right) \vx_{is} \dmP_{kt}\\
        &  +\frac{1}{2} \left( \sum_{i} H_{ki} \vx_{it}^2 \right) \dmP_{kt}^2
\end{split}
\end{equation}
The first order term of this equation is exactly the update rule in Algorithm~\ref{alg:cd-group}. Using $\dmP_{k,:}^*$ to denote the change on the $k$th row after carrying out the update, we arrive at the following inequality
\begin{equation}
\begin{split}
    L(\dmP_{k,:}^*) + \Omega(\mP_{k,:}  + \dmP_{k,:}^*)
    \leq &\upL(\dmP_{k,:}^*) + \Omega(\mP_{k,:}  + \dmP_{k,:}^*) \\
    \leq& \upL( \vzero ) + \Omega(\mP_{k,:}+\vzero) \\
    =& L( \vzero ) + \Omega(\mP_{k,:}+\vzero)
\end{split}
\label{eq:17}
\end{equation}
Note that we start from $\dmP_{k,:}=\vzero$, and we have $\dmP_{k,:}=\dmP_{k,:}^*$ after the update. The inequality in Equation~(\ref{eq:17}) shows that the original loss function decreases after each round of update, and hence this proves the convergence of Algorithm~\ref{alg:cd-group} for any differentiable convex loss function.

\section{Parallel and Efficient Algorithm for Learning to Match}\label{sec:parallel-update}
We propose a parallel and efficient learning-to-match algorithm (\textsc{PL2M}) to further improve the scalability and efficiency by deriving an adaptive estimation of the conflicts caused by parallel updates. Specifically, we consider parallelizing and accelerating Algorithm~\ref{alg:cd-group}. The statistics calculation and preprocessing steps in Algorithm~\ref{alg:cd-group} can be naturally separated into several independent tasks and thus fully parallelized. However, there is strong dependency within $\mP$ update steps, making the parallelization of it a difficult task. We will discuss how we solve the problem next.

\subsection{Parallelization}
Let $S$ be a set of feature indices to be updated in parallel. Assume that the statistics of $G_{ki}$ is up to date as in Algorithm~\ref{alg:cd-group} and we want to change $\mP_{ks}$ for $s \in S$ in parallel. For simplicity of notation, we use $\dmP_{k,:}$ to represent the change in $\mP_{k,:}$. The value of $\upL$ after this change will be
\begin{equation}\label{eq:boundPUPDATE}
\begin{split}
\upL( \dmP_{k,:} ) =& \sum_{i,j\in \Idx} l( \hmY_{ij}, \mY_{ij} ) \\
                    & + \sum_{s\in S}( \sum_{i} G_{ki} \vx_{is}\dmP_{ks} + \frac{1}{2} \sum_{i} H_{ki} \vx^2_{is} \dmP^2_{ks}  ) \\
                    & + \sum_{s\in S}\sum_{t\in S, t\neq s } \sum_{i} H_{ki}\vx_{it} \vx_{is}  \dmP_{kt} \dmP_{ks}
\end{split}
\end{equation}
In a specific case in which $ \vx_{it} \vx_{is}= 0$ for $s\neq t; s,t \in S$, the third line in Equation~(\ref{eq:boundPUPDATE}) becomes zero. This means that the features in the selected set do not appear in the same instance. In such case, the loss can be separated into $|S|$ independent parts and the original update rule can be applied in parallel. Not surprisingly, such condition does not hold in many real world scenarios. We need to remove these troublesome cross terms in the second line, by deriving an adaptive estimation of the conflicts caused by parallel updates, more specifically, by the inequality:
\begin{equation}
    \dmP_{kt} \dmP_{ks} \vx_{it} \vx_{is} \leq \frac{1}{2} |\vx_{it} \vx_{is}|\left(\dmP_{kt}^2 +\dmP_{ks}^2\right)
\end{equation}
With the inequality, we can bound $\upL$ as follows
\begin{equation}
\begin{split}
\upL( \dmP_{k,:} ) \leq      & c + \sum_{s\in S} ( \sum_{i} G_{ki}   \vx_{is} \dmP_{ks}) \\
                             & + \sum_{s\in S} ( \frac{1}{2} \sum_{i} H_{ki}  |\vx_{is}|(\sum_{t\in S} |\vx_{it}|) \dmP^2_{ks} ) \\
                   \triangleq& \upPL(\dmP_{k,:}).
\end{split}
\end{equation}

\SetAlgoSkip{}
\begin{algorithm}[t]
    \caption{Parallel Algorithm for Learning to Match}\label{alg:pcd}
    $\mbox{randomly initialize }\mP,\mQ$\\
    $\mU \leftarrow \mP \vx$, $\mV \leftarrow \mQ \vz$,  $\hmY \leftarrow \mU^\trs \mV$ \{calculate $\mU$, $\mV$, $\hmY$\}\\
    \While{not converge}{
        schedule a partition $P$ of $\{1,2,\cdots, n \}$ \\
        \For{$k=1$ {\bfseries to} $d$}{
            \For{$i=1$ {\bfseries to} $q$ in parallel}{
                $G_{ki} \leftarrow \sum_{j \in \Idx_i }  g_{ij} \mV_{kj}$\\
                $H_{ki} \leftarrow \beta \sum_{j \in \Idx_i}\mV^2_{kj}$\\
            }
            \For{each index set $S \in P$}{
                \For{$i= 1$ {\bfseries to} $q$ in parallel} {
                    $C_{ki} \leftarrow \sum_{s\in S} |\vx_{is}|$\\
                }
                \For{$s \in S$ in parallel}{
                    $x \leftarrow \sum_{i} G_{ki} \vx_{is}$ \\
                    $y \leftarrow \sum_{i} H_{ki} |\vx_{is}| C_{ki}$ \\
                    $\dmP_{ks} \leftarrow  T ( x, y, \mP_{ks}, \alpha, \lambda )$\\
                    $\mP_{ks} \leftarrow \mP_{ks} + \dmP_{ks} $\\
                }
                \For{$i= 1$ {\bfseries to} $q$ in parallel} {
                    $G_{ki} \leftarrow G_{ki} +   \sum_{s\in S} \vx_{is}\dmP_{ks} H_{ki}$ \\
                }
            }
        }
        $\mU = \mP \vx$ \{recalculate buffered $\mU$\}\\
        update $\mQ$ in the same way as $\mP$\\
    }
\end{algorithm}
Obviously this new upper bound $\upPL(\dmP_{k,:})$ can be separated into $|S|$ independent parts and optimized in parallel. Moreover, the sum $\sum_{t\in S} |\vx_{it}|$ is common for all features in set $S$ and is only needed to be calculated once. With this result, we give a parallel and efficient algorithm for learning-to-match, shown in Algorithm~\ref{alg:pcd}.

\subsection{Convergence of Algorithm}\label{sec:pcd-converge}
The relaxation of $\upL(\dmP_{k,:})$ into $\upPL(\dmP_{k,:})$ is performed iteratively in the optimization, and it still attempts to optimize the original objective as in Equation~(\ref{eq:obj}), which is a case much analogous to Expectation-Maximization algorithm in finding a maximum-likelihood solution.
Let $\dmP_{k,:}^*$ be the change in $\mP_{k,:}$ after each parallel update. Since each parallel update optimizes $\upPL(\dmP_{k,:})$, we have the following inequality
\begin{equation}
\begin{split}
    &\upL(\dmP_{k,:}^*) + \Omega(\mP_{k,:}  + \dmP_{k,:}^*) \\
    \leq &\upPL(\dmP_{k,:}^*) + \Omega(\mP_{k,:}  + \dmP_{k,:}^*) \\
    \leq& \upPL( \vzero ) + \Omega(\mP_{k,:}+\vzero) \\
    =& \upL( \vzero ) + \Omega(\mP_{k,:}+\vzero)
\end{split}
\end{equation}
It indicates that $\upL$ decreases after each parallel update. It then follows that the parallel procedure for optimizing the original loss function in Algorithm~\ref{alg:pcd} always converges.

The update rule depends on the statistics $C_{ki}=\sum_{t\in S} |\vx_{it}|$. With the following notation
\begin{equation}
\eta_{ks} = \frac{\sum_{i}H_{ki}\vx_{is}^2 + \lambda}{\sum_{i}H_{ki}C_{ki}|\vx_{is}| + \lambda}
\end{equation}
It can be shown that the parallel update of $\dmP_{ks}$ is shrunken by $\eta_{ks}$ compared to sequential update. Intuitively $\eta_{ks}$ depends on the co-occurrence between features $s\in S$. When features in $S$ rarely co-occur, $\eta_{ks}$ will be close to one, which means that we can update ``aggressively''. When features in $S$ co-occur frequently, $\eta_{ks}$ will get small and we need to update more ``conservatively''. In an extreme case in which no feature co-occurs with each other, $\eta_{ks}=1$ and we get perfect parallelization without any loss of update efficiency. In another extreme case in which we have $|S|$ duplicated features ( $\vx_{is} = \vx_{it} \mbox{ for all } s,t\in S$ ), $\eta_{ks}=\frac{1}{|S|}$, which is extremely conservative given the size of $S$. The advantage of our algorithm is that it \emph{automatically adjusts} its ``level of conservativeness'' by the condition in learning, and thus it always ensures the convergence of the algorithm regardless of the number of threads and the nature of dataset.

The changes in loss function can be analyzed accordingly. Let us consider the simple case in which $\alpha=0$ and only $l_2$ regularization is involved. The change of loss after parallel update can be bounded by
\begin{equation}
\begin{split}
    &\upL(\dmP_{k,:}^*) + \Omega(\mP_{k,:}  + \dmP_{k,:}^*) -  \upL(\vzero) - \Omega(\mP_{k,:}  +\vzero) \\
    \leq & -\frac{1}{4} \sum_{s\in S} \frac{\left(\sum_{i} G_{ki}\vx_{is} \right)^2}{ \sum_{i} H_{ki}  |\vx_{is}|C_{ki} +\lambda } \\
    = & -\frac{1}{4} \sum_{s\in S}\eta_{ks} \frac{\left(\sum_{i} G_{ki}  \vx_{is} \right)^2}{ \sum_{i} H_{ki} \vx_{is}^2+\lambda}\\
\end{split}
\end{equation}
As this inequality indicates, compared to the ideal case in which features do not co-occur, each parallel update's contribution to the loss change is scaled by $\eta_{ks}$. The above analysis also intuitively justifies that $\eta_{ks}$ controls the efficiency of the update.

\subsection{Time Complexity}\label{sec:timeComplextiy}

The time complexity of the efficient algorithm (Algorithm~\ref{alg:cd-group}) is only of $O( d (\|\vx\|_0  + \|\vz\|_0 + |\Idx|) )$. It is linear to numbers of nonzero entries of feature matrices and number of observed entries of $\mY$. Recall the time complexity of the coordinate descent algorithm (Algorithm~\ref{alg:cd-old}), which is $O( d |\Idx|( \frac{\|\vx\|_0}{q}  + \frac{\|\vz\|_0}{p} ) )$.

The \emph{speedup} on $\mP$ updates in Algorithm~\ref{alg:cd-group} is as follows:
\begin{equation}
    \frac{O(d |\Idx| \frac{\|\vx\|_0}{q})}{O(d \|\vx\|_0)} = O(\frac{|\Idx|}{q})
\end{equation}
This corresponds to average number of observed target instances per query instance. Similarly, on the $\mQ$ updates, the \emph{speedup} is about $O(\frac{|\Idx|}{p})$ times. Therefore, the overall \emph{speedup} of Algorithm~\ref{alg:cd-group} over Algorithm~\ref{alg:cd-old} is at least,
\begin{equation}
\begin{split}
\mbox{speedup} = & \frac{O( d |\Idx|( \frac{\|\vx\|_0}{q}  + \frac{\|\vz\|_0}{p} ) )}{O( d (\|\vx\|_0  + \|\vz\|_0 + |\Idx|) )} \\
\geq &  min\{O(\frac{|\Idx|}{q}), O(\frac{|\Idx|}{p})\}\\
= & O(|\Idx| / (q + p))
\end{split}
\end{equation}

In application tasks, this can be at level of $10$ to $100$ such as collaborative filtering and link prediction. When $\|\vx\|_0 + \|\vz\|_0$ is close to~(or smaller than) $|\Idx|$~(datasets like Yahoo!~Music, Tencent Weibo and Movielens-10M), our algorithm runs as fast as the algorithm of plain matrix factorization even though it uses extra features.

For the complexity of the parallel and efficient learning-to-match algorithm (PL2M) described in Algorithm~\ref{alg:pcd}, using $K$ threads to run the algorithm, the computation cost for one round update is $O( \frac{1}{K} d(\|\vx\|_0  + \|\vz\|_0 + |\Idx| ) )$. It is due to the fact that all parts of the algorithm are parallelized. This analysis does not consider the synchronization cost.
In real world settings, we need to take synchronization cost into consideration, the corresponding time complexity becomes $O( \frac{1}{K} d((\|\vx\|_0  + \|\vz\|_0)(1+\sigma) + |\Idx| ) )$, where $\sigma$ denotes variance of computation costs by parallel tasks. Assume that we have $K$ tasks and the time costs of the tasks are $T_1, T_2,\cdots T_{K}$. We define $\sigma = max_i( T_i )  / Mean_i( T_i )$, since the training is delayed by the slowest task. To achieve maximum speedup, we need to schedule the tasks well such that the load of each task is average, which is always feasible when $|S|$, $p$, and $q$ are large. Therefore, our algorithm can gain almost $K$ times speedup.

In real world applications, there is a trade-off between the size of \emph{parallel coordinate set} $|S|$ and the parameter $\sigma$, especially when different features have different levels of sparsity in the dataset. When we increase the size of \emph{parallel coordinate set} $|S|$, we can divide the task into $K$ threads in a more balanced way. On the other hand, $\eta_{ks}$ will decrease as we increase $|S|$, making the update more conservative. Thus a parallel coordinate set needs to be chosen to balance convergence and acceleration. In fact, we need to empirically choose $S$ such that each instance is covered by only a few nonzero features and the task size is large enough to run in a fairly balanced way.

In this paper, we fix $|S|$ and randomly partition elements from the feature indices to generate a set of disjoint subsets $S$ in each round. We note that there can be more sophisticated scheduling strategies to select $S$, which is beyond the scope of this paper and can be an interesting topic for future research.

\begin{table*}[t]
\centering
\caption{Details of 4 Datasets} \label{tbl:dataset}
\begin{tabular}{|l|c|c|c|c|l|l|}\hline
    Dataset          & $q \times p$ & $\|\vx\|_0$ & $\|\vz\|_0$ & $|\Idx|$ & Task & Available Features\\    \hline
    Yahoo! Music     & $1M \times 0.6M$  & $263M$ & $1M$  & $250M$& Collaborative Filtering & User Feedback~\cite{Koren:KDD08}, Taxonomy\\\hline
    Tencent Weibo    & $2M \times 5K$   & $55M$ & $83K$ & $93M$& Social Link Prediction & Social Network, User Profile, \\
                     &                  &          &      &      &                      & Taxonomy, Tag\\\hline
    Flickr           & $0.65M \times 20K$  & $451M$ & $20K$ & $39M$& Image Tagging & MAP, Sift Descriptors of Image\\\hline
    Movielens-10M     & $69K \times 10K$ & $10M$ & $10K$ & $9M$ & Collaborative Filtering & User Feedback~\cite{Koren:KDD08} \\ \hline
\end{tabular}
\end{table*}

\section{Related Work}\label{sec:rel}

MF models~\cite{Koren:MF09} are arguably the most successful approach to learning-to-match. They have been applied to a wide range of real world problems, especially FMF models, which achieve state-of-the-art results, outperforming the models of MF in many different tasks, with different types of features used. In collaborative filtering, user feedback information (SVD++)~\cite{Koren:KDD08}, user attribute information~\cite{agarwal:kdd}, and product attribute information~\cite{stern:www} are incorporated into models to further enhance the accuracies in prediction. In web search~\cite{Wu:WSDM13, wu2013learning}, term vectors of queries and documents are utilized as features to significantly improve relevance ranking. FMF models also give the best results in link prediction in KDD Cup 2012~\cite{Menon:ECML11, Chen:SIGIR2012, Chen:KDDCUP}. These works demonstrate the effectiveness of the learning-to-match models, but also create necessity for parallelization of the learning algorithms.

There has been much effort on parallelizing the process of plain matrix factorization. For example, Gemulla et al.~\cite{Gemulla:KDD11} propose a method of distributed stochastic gradient descent for MF. Yu et al.~\cite{Yu:ICDM12} introduce a parallel coordinate descent algorithm for MF. An alternating least square method is proposed for MF as well~\cite{Zhou:aaim08}. Recently, Zhuang et al.~\cite{Zhuang:Recsys2013} improve the efficiency of parallel stochastic gradient descent for MF by making a better scheduling of updates. Liu et al.~\cite{dnmf} propose a distributed algorithm for nonnegative matrix factorization for web dyadic data analysis. The method of Probabilistic Latent Semantic Indexing is parallelized for Google news recommendation~\cite{googlenews}. However, all the models on parallelizing plain matrix factorization replies on the fact that the rows and columns can be naturally separated and the parameters can be independently updated, and therefore cannot work on FMF due to the complex feature dependencies in the updating steps.

There is only a little work focusing on acceleration of coordinate descent for FMF. The most recent one scales up coordinate descent by making use of repeating patterns of features~\cite{rendle2013scaling}. However, it is specialized for the least-squares loss and probit loss. Although the idea of avoiding repeated calculations is similar, our algorithm takes a completely different approach and can handle any general convex loss functions. Other algorithms of parallel coordinate descent, such as~\cite{collins2002logistic}~and~\cite{mukherjee2013parallel} cannot be directly applied to FMF, because it is difficult for them to handle complex feature dependencies in FMF.

As general parallelization technique, the Hogwild! algorithm~\cite{hogwild} can be applied to our problem. However, its time complexity is $O( d |\Idx|( \frac{\|\vx\|_0}{q}  + \frac{\|\vz\|_0}{p} ) )$, the same as Algorithm~\ref{alg:cd-old}, due to the repeated calculations. Using the same number of threads, as analyzed in time complexity sections, it is theoretically $O(|\Idx| / (q + p))$ times slower than our parallel algorithm. In experiments, our parallel algorithm runs averagely about 5 times faster than Hogwild!. Another thread of related work is parallelization of coordinate descent algorithms. There have been studies on parallelizing coordinate descent for linear regression~\cite{Bradley:ICML2011,Scherrer:ICML2012,Scherrer:NIPS2012}, other than matrix factorization~\cite{Yu:ICDM12}. The convergence of these algorithms depends on the spectrum of covariance matrix, which changes in each round in our learning setting~(due to the changes in $\mU$ and $\mV$), and thus the algorithms cannot be directly applied to our problem. Our algorithm makes use of parallel update to minimize an upper bound re-estimated each round to ensure convergence, which can also be viewed as a kind of minorization-maximization algorithm~\cite{Lange:MM}.

\section{Experiments}\label{sec:exp}
In this section, we introduce our experimental results on several matching tasks using benchmark datasets. We first conduct comparison on accuracies between feature-based matrix factorization and plain matrix factorization. We then make comparisons on accuracies and efficiencies between our method of parallel learning-to-match and the baselines, including Hogwild!~\cite{hogwild}. Finally, we conduct analysis on the efficiency of our parallel learning algorithm.

\subsection{Datasets}

Four datasets representing different types of learning-to-match tasks are chosen. Details of the datasets are summarized in Table~\ref{tbl:dataset}.

The first dataset is Yahoo!~Music~Track1~\footnote{http://kddcup.yahoo.com/datasets.php} from the Yahoo!~Music website.
The dataset is among the largest public datasets for collaborative filtering. We use the official split of the dataset for experiments.
As features, we use the implicit feedback of users~\cite{Koren:KDD08} as well as the taxonomical information between the tracks, albums and artists, in addition to the indicators of users and tracks. Because it is an item rating dataset, we choose square loss as the loss function and use Root Mean Square Error~(RMSE) as evaluation measure.

The second dataset is Tencent Weibo~(microblog)\footnote{http://kddcup2012.org/c/kddcup2012-track1/data}, for social link prediction. The task is to predict a potential list of celebrities that a user will follow.
The dataset is split into training and test data by time, with the test data further split into public and private sets for independent evaluations.
We use the training set for learning and the public test set for evaluation. We use logistic loss as the loss function and MAP@K as evaluation metric, which is officially adopted in the KDD Cup competition~\footnote{http://kddcup2012.org/c/kddcup2012-track1/details/Evaluation}.
The matrix data is extremely sparse, with only on average two positive links per user. Furthermore, about $70\%$ of users in the test set have no following records in the training set.
However, there are lots of additional information available, including social network and interaction (i.e., retweeting and commenting) records, profiles of users, categories of celebrities, and tags/keywords of users.
The information is used as features of the task.

The third dataset is for automatic annotating images crawled from Flickr\footnote{http://www.flickr.com}.
The dataset contains $0.65$ million images and each image is associated with on average four tags.
We select the $20,000$ most frequently occurring tags as the tag set.
We randomly select $10,000$ images as test set and use the rest of images as training set.
We use the bag-of-words vector of SIFT descriptors as features for images, and indicator vectors as features for tags.
Logistic loss is chosen as the loss function.
In testing, we generate a rank list of tags and use P@K(Precision at K) and MAP as evaluation metrics.

The fourth dataset is also for collaborative filtering, provided by Movielens\footnote{http://www.movielens.org/}. We use the official split of dataset for experiments. This dataset is added because Hogwild! cannot run on Yahoo! Music dataset due to its high time complexity. In addition to the indicators of users and movies, the implicit feedbacks of users~\cite{Koren:KDD08} are used as features. Similar to Yahoo!~Music dataset, we choose square loss as the loss function and RMSE as evaluation metric.

\subsection{Experiment Setting}
We have implemented our parallel and efficient algorithm for learning-to-match (\textsc{PL2M}) using OpenMP\footnote{http://www.openmp.org}.
The experiments are conducted on a machine with an Intel Xeon CPU E5-2680~(8 cores, supporting 16 threads at 2.70GHz, 128GB memory).
We utilize up to 15 working threads and reserve one thread for scheduling.

We compare the performance of PL2M with those of serial algorithm for learning-to-match algorithm (denoted as L2M) and the Hogwild! algorithm~\cite{hogwild}.
To simplify the notations, here we use \PCD{$T$} to refer to the parallel algorithm for learning-to-match with parallel set $|S|=T$~(e.g \PCD{5K} means the parallel algorithm with $|S|=5,000$).
Hogwild! is the only one that can be directly applied to our problem as mentioned in Section~\ref{sec:rel}. We have also implemented Hogwild! using OpenMP.
\emph{All matrix operations} mentioned in the algorithms take the advantages of data sparsity. PL2M, \SCD, and Hogwild! share the same codes of elementary operations.

We empirically set $\lambda=1, \alpha=0.1$ and $d=64$ for \SCD and PL2M through all our experiments.
To make fair comparison, the parameters of Hogwild! including learning rate, $\lambda$ and $\alpha$ are tuned with cross validation on training set.

\begin{figure*}[t]
  \centering
  \subfigure[TLC on Yahoo! Music]{
    \includegraphics[scale=0.4]{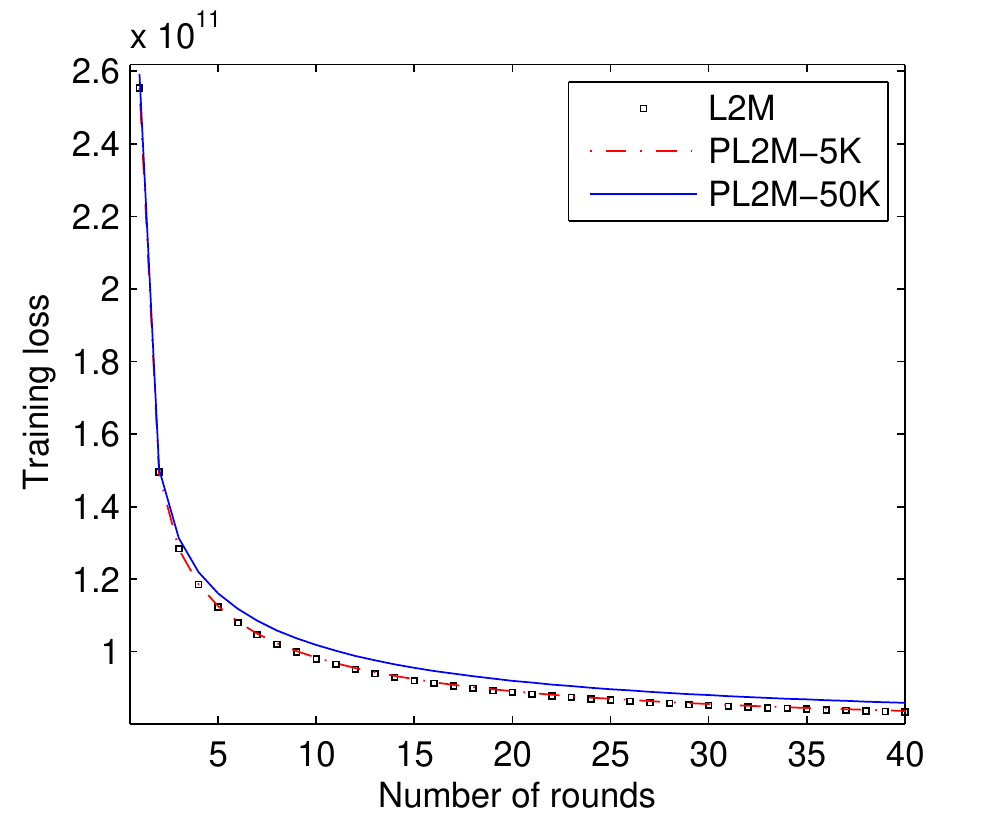}
  }
  \subfigure[TLC on Tencent Weibo]{
    \includegraphics[scale=0.4]{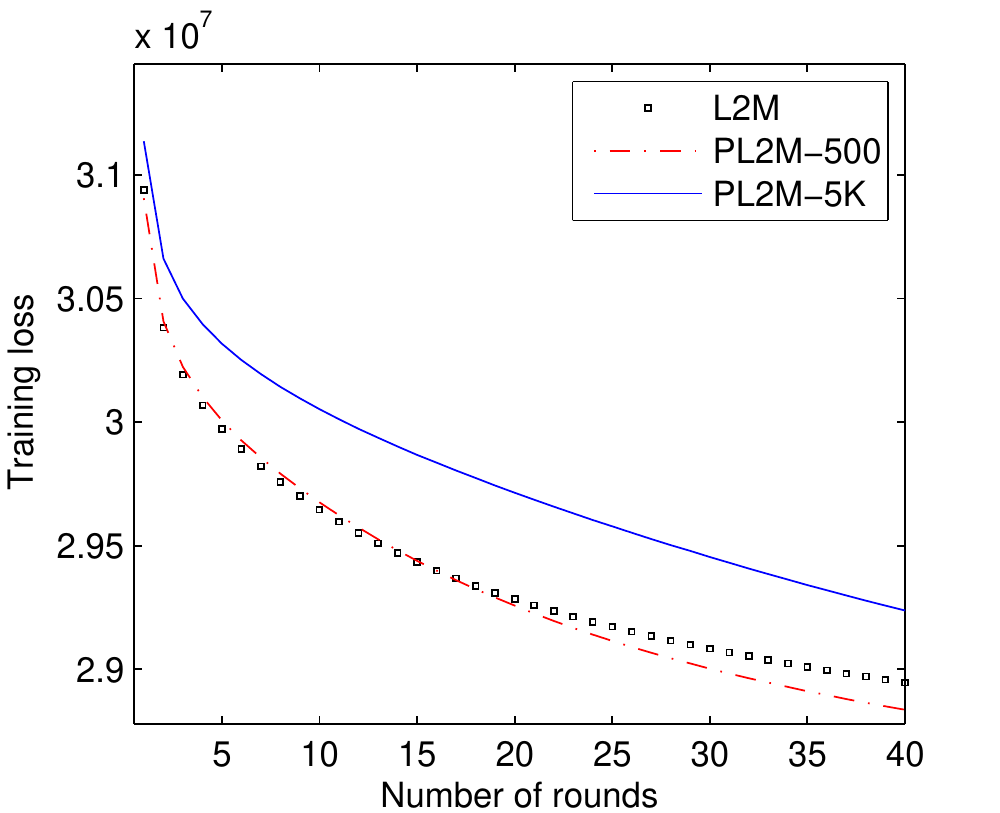}
  }
  \subfigure[TLC on Flickr]{
    \includegraphics[scale=0.4]{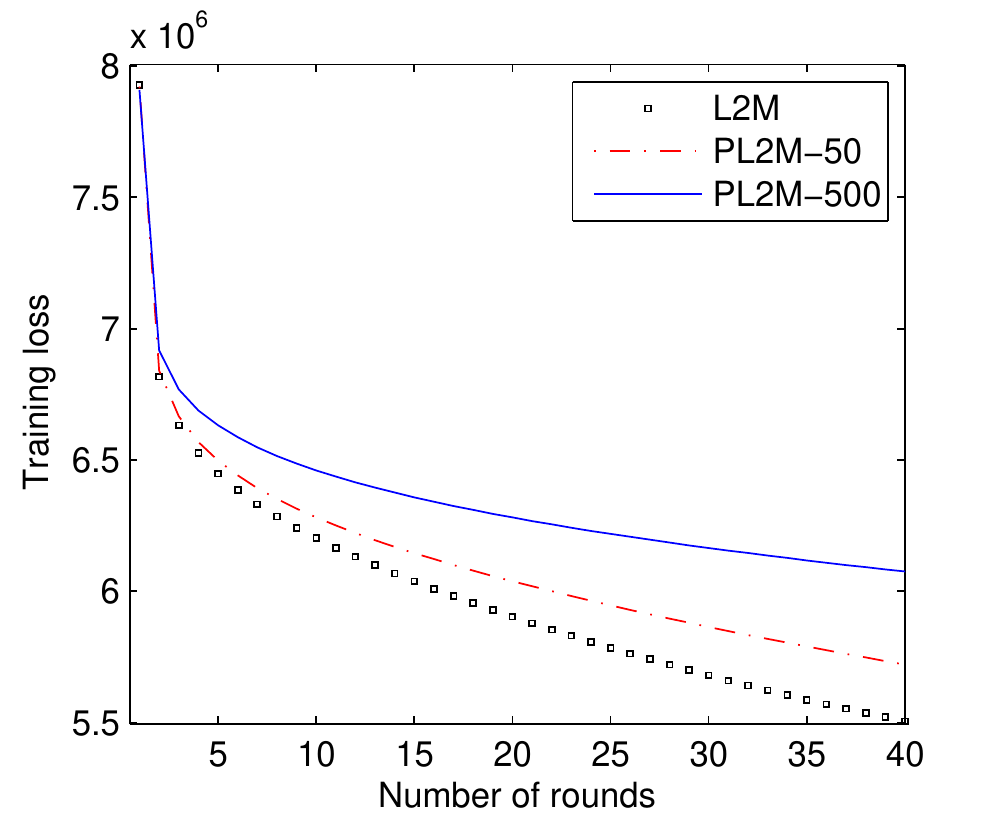}
  }
  \subfigure[TLC on Movielens-10M]{
    \includegraphics[scale=0.4]{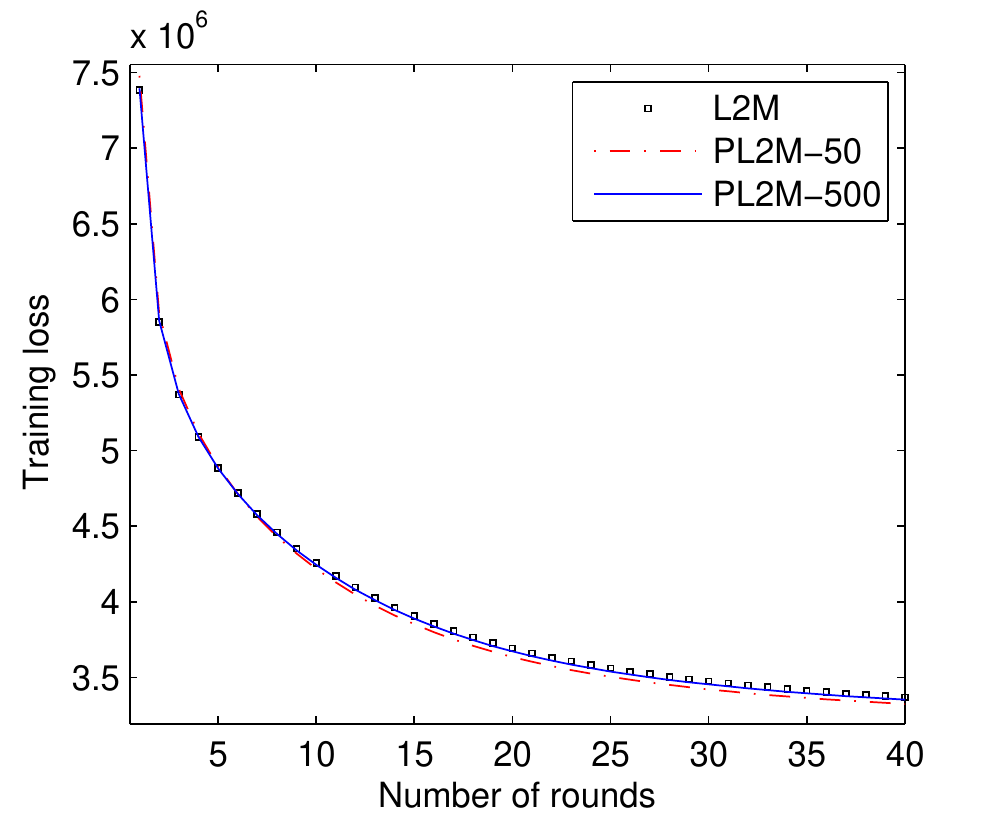}
  }
  \caption{Training Loss Convergence(TLC) on Four Datasets}
  \label{fig:TLC}
\end{figure*}

\subsection{Usefulness of Features}

\begin{figure}[t]
\center
    \subfigure[on Yahoo! Music] {
        \includegraphics[scale=0.35]{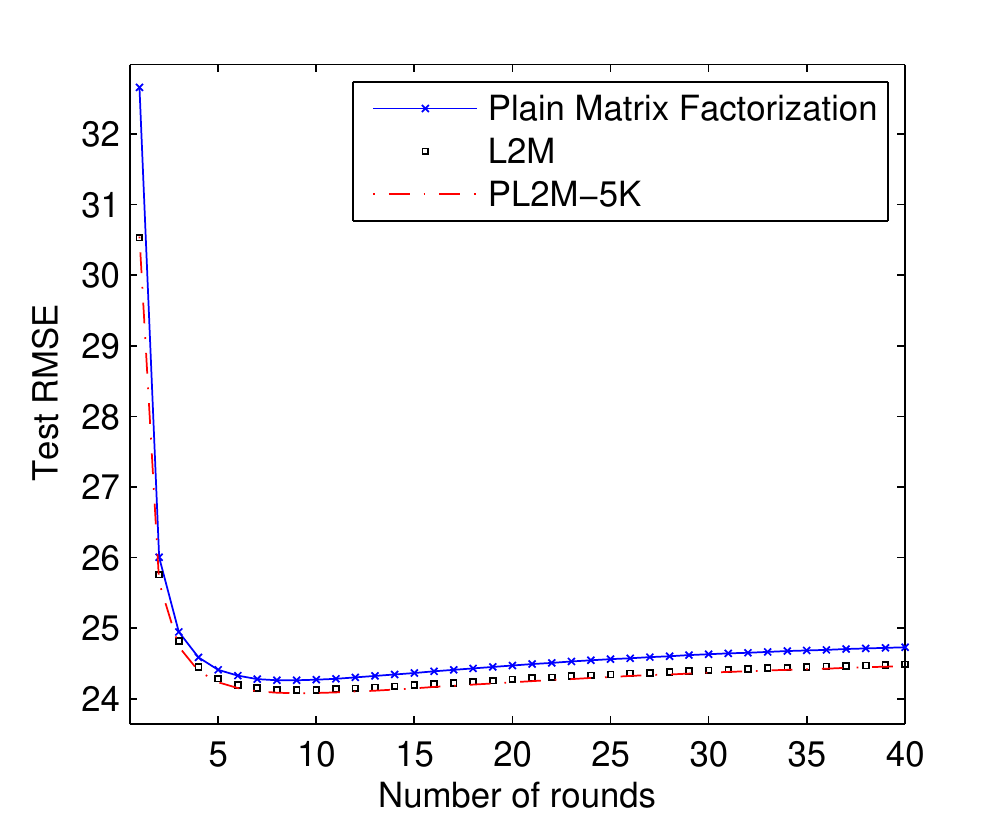}
        \label{fig:yahoo-rmse}
    }
    \subfigure[on ML-10M] {
        \includegraphics[scale=0.35]{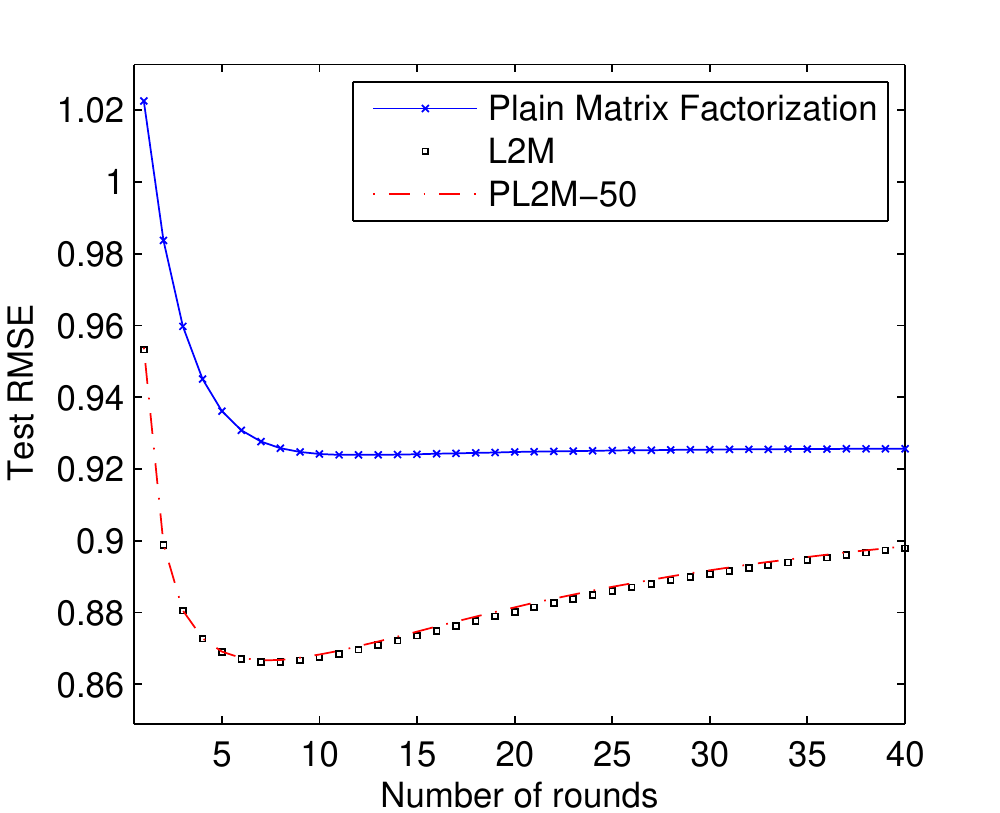}
        \label{fig:ml10m-rmse}
    }
    \caption{Test RMSE}
    \vspace{-0.25in}
\end{figure}

We make comparison between FMF and MF to investigate the effectiveness of the features.
We first compare FMF (by the algorithm of L2M) and MF in terms of test RMSE on the Yahoo! Music dataset in Figure~\ref{fig:yahoo-rmse}. From the result, we can find that the FMF model converges faster and achieves better results than the MF model. This result is consistent with the result reported in~\cite{Koren:KDD08,rendle:tist2012} and confirms the importance of using features in this problem. The test error first decreases but increases again in different rounds of training, indicating that training can stop at about 5 rounds.

The results of Tencent Weibo dataset are shown in Table~\ref{tbl:tencentMAP}.
Since MF gives similar performance as the popularity based algorithm that only considers the popularity of each target node, and thus the result is not reported.
Here the suffix ALL stands for the FMF model using all the available features shown in Table~\ref{tbl:dataset}.
We also evaluate the performance of the FMF model with only social network information, with suffix SNS.
From Table~\ref{tbl:tencentMAP}, we can see that this dataset is extremely biased toward popular nodes.
However, it is still possible to improve the results using social network information, and the auxiliary features can help to achieve the best performance.
Note that \PCD{500}-ALL has achieved the best result in Tencent Weibo dataset (actually our method is the same as the champion system on this dataset~\cite{Chen:KDDCUP}).

\begin{table}[t]
\centering
\caption{Results of Social Link Prediction~(Tencent Weibo) in MAP} \label{tbl:tencentMAP}
\begin{tabular}{|l|c|c|c|}\hline
    Setting         & MAP@1    & MAP@3     & MAP@5 \\ \hline
    Popularity        & 22.54\%  & 34.65\%   & 38.28\%  \\\hline
    \SCD[-SNS]          & 24.10\%  & 36.56\%   & 40.19\%  \\\hline
    \PCD{500}-SNS    & 24.18\%  & 36.65\%   & 40.27\% \\\hline
    \SCD[-ALL]          & 25.44\%  & 38.02\%   & 41.63\% \\\hline
    \PCD{500}-ALL & 25.52\%  & 38.14\%   & 41.75\%  \\\hline
\end{tabular}
\vspace{-0.25in}
\end{table}

The performance on Flickr test set is shown in Table~\ref{tbl:FlickrMAP}.
Because training and test images do not overlap, we cannot use MF to make prediction, and thus we adopt the use of popularity scores as a baseline.
From the result, we can find that FMF can improve upon the popularity method, and assign relevant tags using image content features.

\begin{table}[t]
\centering
\caption{Results of Image Tagging~(Flickr) in Precision} \label{tbl:FlickrMAP}
\begin{tabular}{|l|c|c|c|}\hline
    Setting         & MAP      & P@1     & P@3      \\\hline
    Popularity        & 3.96\%   & 4.63\%  & 4.08\%  \\\hline
    \SCD            & 7.18\%   & 11.05\% & 8.76\%    \\\hline
    \PCD{50}        & 7.59\%   & 11.86\% & 9.19\%   \\\hline
\end{tabular}
\end{table}

The test RMSE curves of different algorithms on Movielens-10M dataset are shown in Figure~\ref{fig:ml10m-rmse}. From the result, we can find that the FMF model converges faster and achieves better results than the MF model. This result demonstrates the importance of using features in this problem. The test error first decreases but increases again in different rounds of training, which indicates that training can be stopped at about 7 rounds.

\begin{figure*}[t]
\centering
 \subfigure[on Yahoo!~Music]{
    \includegraphics[scale=0.5]{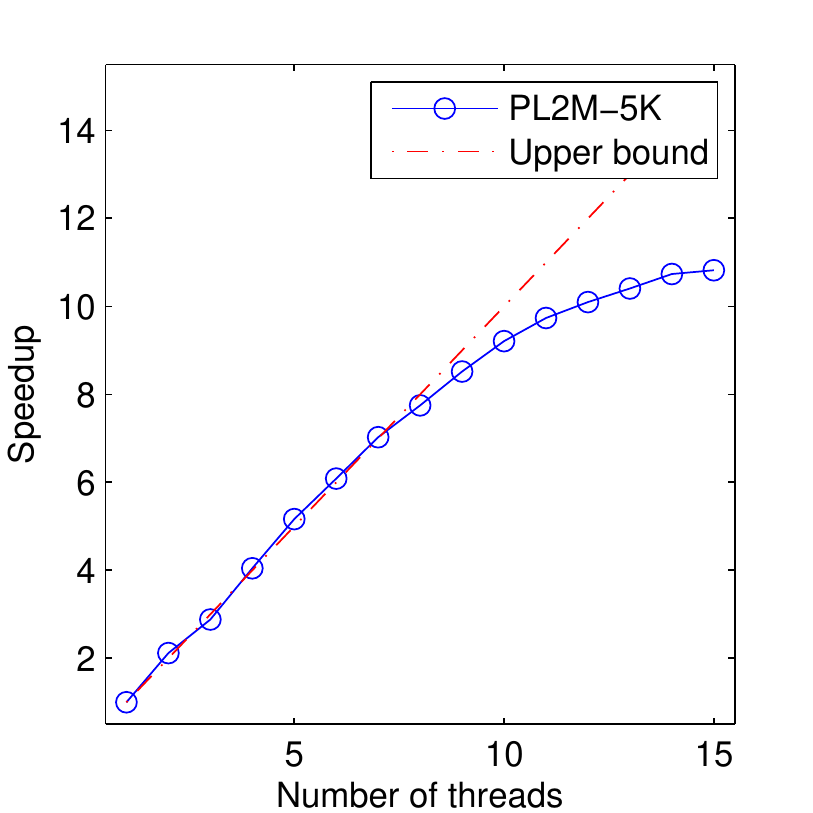}
    \label{fig:yahoo-speedup}
  }
  \subfigure[on Tencent Weibo]{
    \includegraphics[scale=0.5]{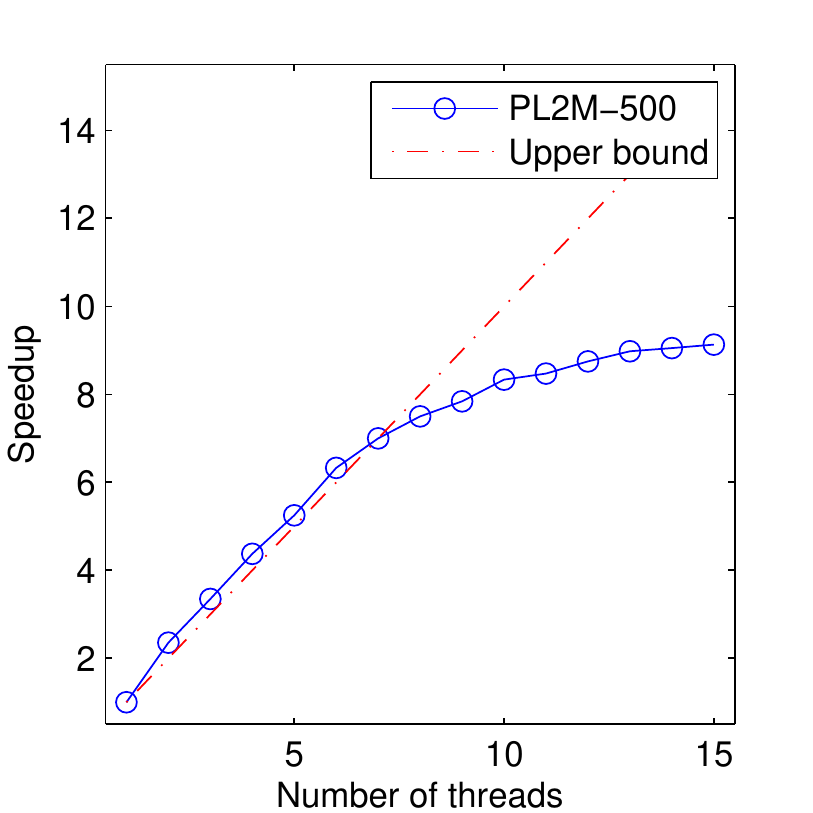}
    \label{fig:tencent-speedup}
  }
  \subfigure[on Flickr]{
    \includegraphics[scale=0.5]{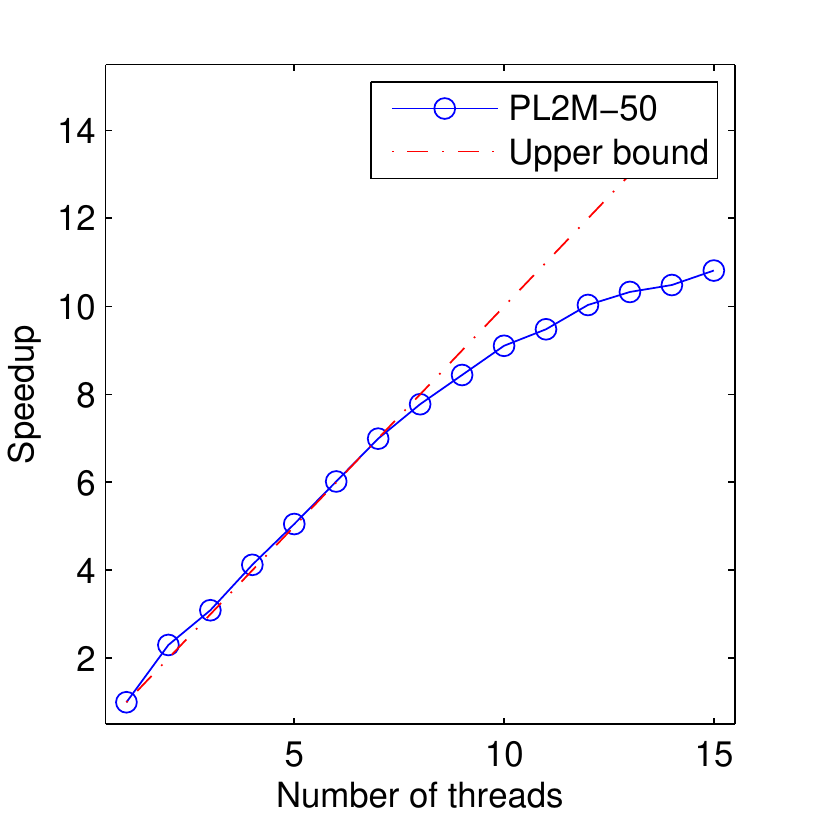}
    \label{fig:flickr-speedup}
  }
  \subfigure[on Movielens-10M]{
    \includegraphics[scale=0.5]{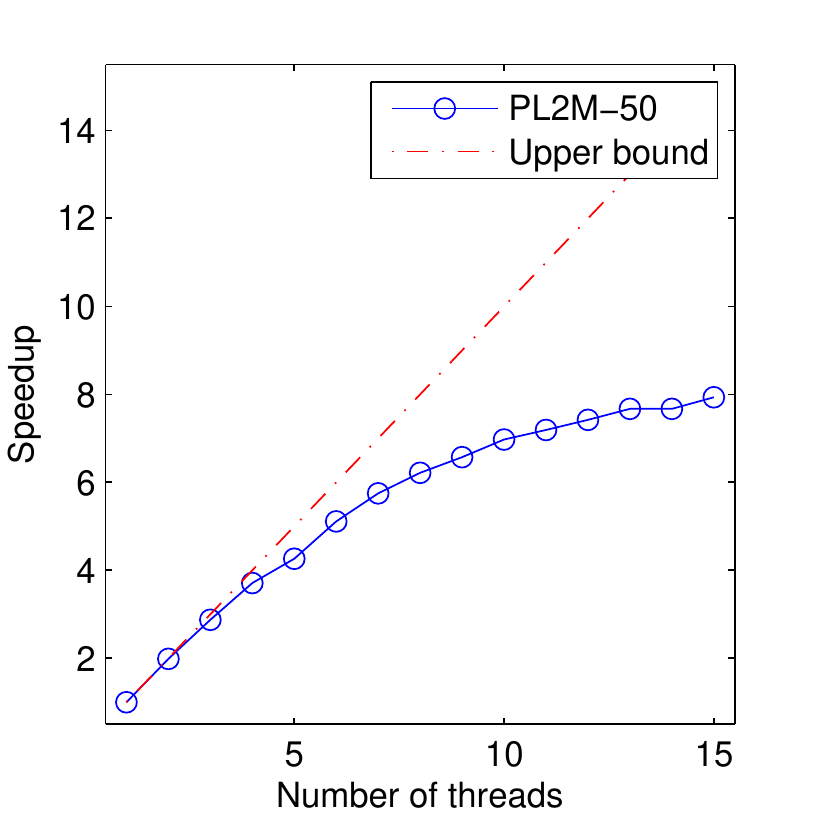}
    \label{fig:flickr-speedup}
  }
  \caption{Efficiency Evaluation on Four Datasets (Speedup Curves)}
  \label{fig:speedup}
  \vspace{-0.25in}
\end{figure*}

\subsection{PL2M versus L2M}
We make comparison between PL2M and L2M in terms of accuracy and efficiency.

As shown in Figure~\ref{fig:yahoo-rmse}, Table~\ref{tbl:tencentMAP}$\sim$\ref{tbl:FlickrMAP}, and Figure~\ref{fig:ml10m-rmse}, PL2M always gives comparable or even better test errors as L2M, indicating that PL2M makes no sacrifice on accuracy in the parallelization.

Figure~\ref{fig:TLC} gives the training loss curves of PL2M and L2M. From the figure, we can observe that PL2M always converges following the lower bound given by L2M at the beginning. This is consistent with our theoretical result on convergence in Section~\ref{sec:parallel-update}. That is, if PL2M and L2M start with the same initial values, PL2M can perform at most as well as L2M.

The things get changed, however, as the training goes on. On Tencent Weibo dataset, \PCD{500} converges slightly better than \SCD after $20$ rounds. On Movielens-10M dataset, although the training loss curves of L2M, \PCD{500}, and \PCD{50} are almost the same, the training loss of PL2M is lower in the end. This may due to the fact that the loss function is non-convex for $\mP$ and $\mQ$. After several updates, for example, 20 rounds, the values of $\mP$ and $\mQ$ are quite different so that the two methods finally converge to different local minimums.

From these figures, we can also observe that \PCD{50K} converges slower than \PCD{5K} in Yahoo!~Music dataset, \PCD{5K} converges slower than \PCD{500} on Tencent Weibo dataset, \PCD{500} converges slower than \PCD{50} on Flickr dateset. \PCD{500} converges a little slower than \PCD{50} on Movielens-10M dateset. These are consistent with our previous theoretical result that smaller $|S|$ leads to faster convergence.

\subsection{PL2M versus Hogwild!}
We make comparison between PL2M and Hogwild!. Since Hogwild! is based on stochastic gradient descent, some parameters such as learning rate need to be tuned. After fine tuning the parameters using cross validation on the training set for Hogwild!, including learning rate, $l_2$ coefficient $\lambda$, and $l_1$ coefficient $\alpha$, we have obtained its performance in  Table ~\ref{tbl:hogwild}, including test error, running time of one round of training, and number of rounds needed to get the best test error. Both Hogwild! and PL2M use $8$ threads.

We can see that the running time for each round of PL2M is much shorter than that of Hogwild!, while their test errors are similar. The difference of running time on Tencent Weibo is not as much as Movielens-10M and Flickr datasets because they have less features. This is consistent with our theoretical result about the time complexity in Section~\ref{sec:ecd} and Section~\ref{sec:parallel-update}.

Furthermore, Hogwild! needs more training than PL2M to achieve its best test errors. For example, Hogwild! needs $20$ rounds but PL2M needs only $7$ rounds to achieve their best test errors in Movielens-10M. Therefore, the total running time for PL2M to get the best performance is much smaller than that of Hogwild!.

\begin{table}[t]
\centering
\caption{PL2M versus Hogwild! (8 threads)}
\label{tbl:hogwild}
\begin{tabular}{|l|c|c|c|c|}\hline
    Dataset          & Method & sec/round & rounds & test error \\    \hline
    Tencent Weibo    &  Hogwild! & 104.3 & 14   & 24.96\%      \\
    (MAP@1)          &  PL2M & 70.1 & 5 & 25.52\%  \\   \hline
    Flickr           &  Hogwild! &  1117.0 & 59    & 7.59\%\\
    (MAP)            & PL2M  &   155.0 & 33 & 7.59\% \\ \hline
    Movielens-10M    &  Hogwild!    & 162.2 & 20    & 0.8756\\
    (RMSE)           &  PL2M    & 18.5 & 7    & 0.8666 \\ \hline
\end{tabular}
\end{table}

\subsection{Scalability of PL2M}
Finally, we evaluate the scalability of the parallel learning-to-match algorithm (PL2M). We test the average running time of \PCD{5K} on Yahoo! Music dataset, \PCD{500} on Tencent Weibo dataset, \PCD{50} on Flickr dataset, \PCD{50} on Movielens-10M with varying numbers of threads and evaluate the improvement in efficiency.

As shown in Figure~\ref{fig:speedup}, the speedup curves are similar on Yahoo!~Music, Tencent Weibo, and Flickr datasets, but the curve converges earlier on the Movielens-10M dataset. This is because that Movielens-10M is relatively smaller than the others and \PCD{50} runs really fast on Movielens-10M, which only needs about 18 seconds when 8 threads are used. Although the speedup gained by parallelization is not as much as that on other datasets, the parallel algorithm can also provide accelerations.

On the first 3 datasets, PL2M can achieve almost linear speedup with less than $8$ threads, but the speedup gain slows down with more threads. We observe that the working threads are still fully occupied with more than 8. We conjecture that this turning point is due to the fact that the number of physical cores of the machine is only 8. From the results, we can find that PL2M is able to gain about $9$ times speedup using $10$ threads, confirming the scalability of the parallel algorithm.

In summary, the speedup gained by the parallel algorithm is significant, and thus it can easily handle hundreds millions of instances and features on a single machine.

\section{Conclusion}\label{sec:con}
We have proposed a parallel and efficient algorithm for learning-to-match, more specifically feature-based matrix factorization, a general and state-of-the-art approach. Our algorithm employs (1) iterative relaxations to solve the conflicts caused by parallel updates, with provable convergence guarantee on minimizing the original objective function, and (2) accelerate the computation by avoiding the repeated calculations caused by features, for any general convex loss functions. As a result, our algorithm can easily handle data with hundreds of millions of objects and features on a single machine. Extensive experimental results show that our algorithm is both effective and efficient when compared to the baselines.

As future work, we plan to (1) extend the algorithm to a distributed setting instead of the current multi-threading, (2) find better scheduling strategies for making parallel updates with a guaranteed bound of speedup, and (3) apply the technique developed in this paper to the parallelization of other learning methods, such as Markov Chain Monte Carlo~(MCMC) learning methods for learning-to-match problem.



\end{document}